\documentclass[runningheads]{llncs}
\usepackage{graphicx}
\usepackage{cite}
\usepackage{array}
\usepackage{amssymb}
\usepackage{bbding}
\usepackage{pifont}
\usepackage{xcolor}
\usepackage{multirow}

\let\oldding\ding
\renewcommand{\ding}[2][1]{\scalebox{#1}{\oldding{#2}}}
\newcolumntype{M}[1]{>{\centering\arraybackslash}m{#1}}

\begin{document}
\title{Melanoma Diagnosis with Spatio-Temporal Feature Learning on Sequential Dermoscopic Images}

\author{Zhen Yu\inst{1, 2}, Jennifer Nguyen\inst{3}, Xiaojun Chang\inst{4}, John Kelly\inst{3}, Catriona Mclean\inst{2}, Lei Zhang\inst{2}, Victoria Mar\inst{3}, Zongyuan Ge \inst{1(}\Envelope\inst{)}}
\authorrunning{**** et al.}
\institute{ Monash eResearch Center, Monash University \and
Central Clinical School, Monash University \and
School of Public Health and Preventive Medicine, Monash University \and
Faculty of Information Technology, Monash University
\\
\email{zongyuan.ge@monash.edu}, \url{https://mmai.group}}
\maketitle

\begin{abstract}
Existing studies for automated melanoma diagnosis are based on single-time point images of lesions. However, melanocytic lesions de facto are progressively evolving and, moreover, benign lesions can progress into malignant melanoma. Ignoring cross-time morphological changes of lesions thus may lead to misdiagnosis in borderline cases. Based on the fact that dermatologists diagnose ambiguous skin lesions by evaluating the dermoscopic changes over time via follow-up examination, in this study, we propose an automated framework for melanoma diagnosis using sequential dermoscopic images. To capture the spatio-temporal characterization of dermoscopic evolution, we construct our model in a two-stream network architecture which capable of simultaneously learning appearance representations of individual lesions while performing temporal reasoning on both raw pixels difference and abstract features difference. We collect 184 cases of serial dermoscopic image data, which consists of histologically confirmed 92 benign lesions and 92 melanoma lesions, to evaluate the effectiveness of the proposed method. Our model achieved AUC of 74.34\%, which is $\sim$8\% higher than that of only using single images and $\sim$6\% higher than the widely used sequence learning model based on LSTM.
\end{abstract}

\section{Introduction}
In practice, dermatologists often diagnose or rule out melanoma by evaluating sequential dermoscopic images of a lesion over time. The rationale behind this is that benign melanocytic naevi will remain fairly stable over time, and lack of change is reassuring~\cite{ref_1, ref_2}. Conversely, melanoma may develop from a pre-existing benign naevus, with changes diagnostic for melanoma difficult to appreciate at a single time-point (shown in Fig. 1). \cite{ref_3} shows 42\% of superficial spreading melanomas (the most common subtype) arise from a pre-existing naevus. 
However, existing computational algorithms for melanoma diagnosis tend to make the decision based on imaging data of lesions from single time-points~\cite{ref_4, ref_5}. This mechanism can be problematic, as ignoring the temporal information of dermoscopic changes may lead to misdiagnosis of borderline lesions. In this study, we 
\begin{figure}
\centering
\includegraphics[scale=0.2]{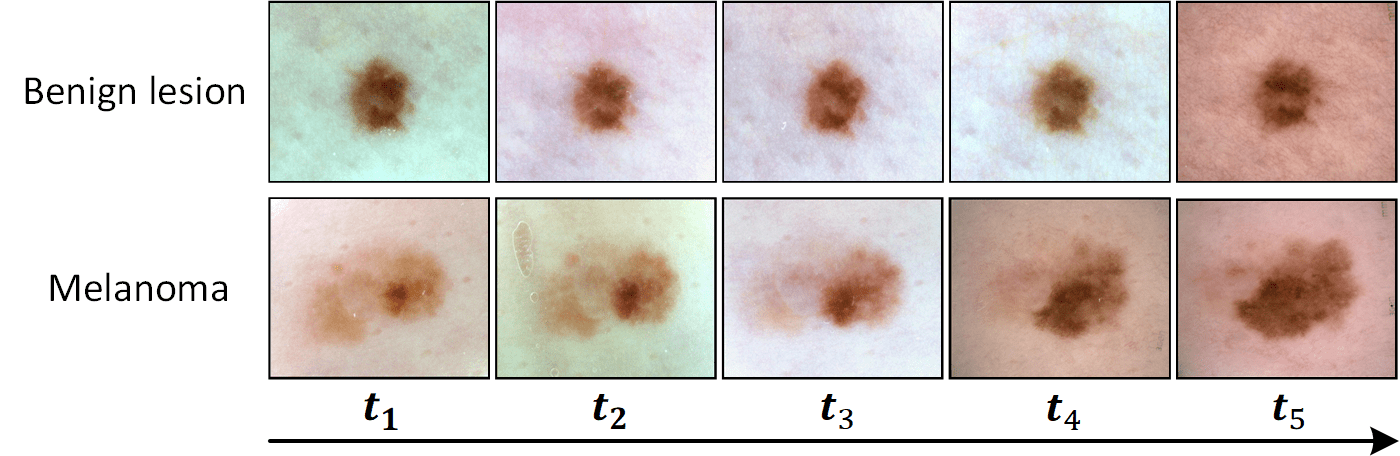}
\caption{Lesions de facto are progressively evolving. The benign lesion remains stable, while the malignant melanoma exhibits substantial focal enlargement. }\label{fig1}
\end{figure}
aim to address the limitation of existing melanoma diagnosis algorithms based on single-time point lesion images and improve the diagnostic accuracy by incorporating temporal dynamics of lesion growth as an extra clue. To the best of our knowledge, this is the first study to exploit modeling spatio-temporal characterization of lesion evolution for automatic melanoma diagnosis.

To effectively learn spatio-temporal features from longitudinal imaging data, there are two main categories of methods. Methods in the first category usually utilise convolutional neural network (CNN) to extract high-level abstract representations from each input image, and then perform temporal aggregation via general pooling or recurrent neural network (RNN)~\cite{ref_6, ref_7}. 
Another line of attempts directly learns spatio-temporal features using 3D networks~\cite{ref_8, ref_9, ref_13}. By stacking multiple images as inputs, these models can effectively extract high level spatio-temporal features. State-of-the-art results were achieved using the above methods in a range of sequential image processing tasks, especially in video analysis domain \cite{ref_8, ref_9}. Nevertheless, in medical domain the number of dermoscopic imaging samples from each patient varies in length and often with much lower number compared to available frames from video sequences (e.g. 3$\sim$5/patient vs 1800/clip). Moreover, subtle changes are not well-captured from the existing video processing methods as their focus on changes from a global scale with both background and foreground considered. Therefore, standard sequential learning methods would not be suitable for a melanoma diagnosis task. 

To effectively track subtle changes in appearance and also capture discriminative spatial features from sequential lesion images for melanoma diagnosis, we propose a framework constructed in a two-stream network manner. The temporal stream conducts temporal reasoning among two consecutive dermoscopic images at both pixel-level and feature-level while the spatial stream focuses on encoding distinguishable appearance abstraction from each individual lesion. We collect 184 serial dermoscopic image data from 184 patients with histologically confirmed diagnosis results and perform experiments to evaluate the effectiveness of the proposed approach. The results show that the model learned from sequential images can achieve significant improvement compared with that of only using snapshot image. Moreover, the proposed method outperforms other widely used sequence learning model for video processing by a large margin.

\section{Method}
Fig. 2 shows the overview of the proposed framework which consists of two main sub-networks: the spatial appearance encoding network and the temporal difference encoding network. The two-stream network capable of simultaneously learning abstract appearance representations from individual lesions while capturing temporal relations among consecutive images from both images difference and multi-level spatial features difference.
\begin{figure}
\centering
\includegraphics[scale=0.058]{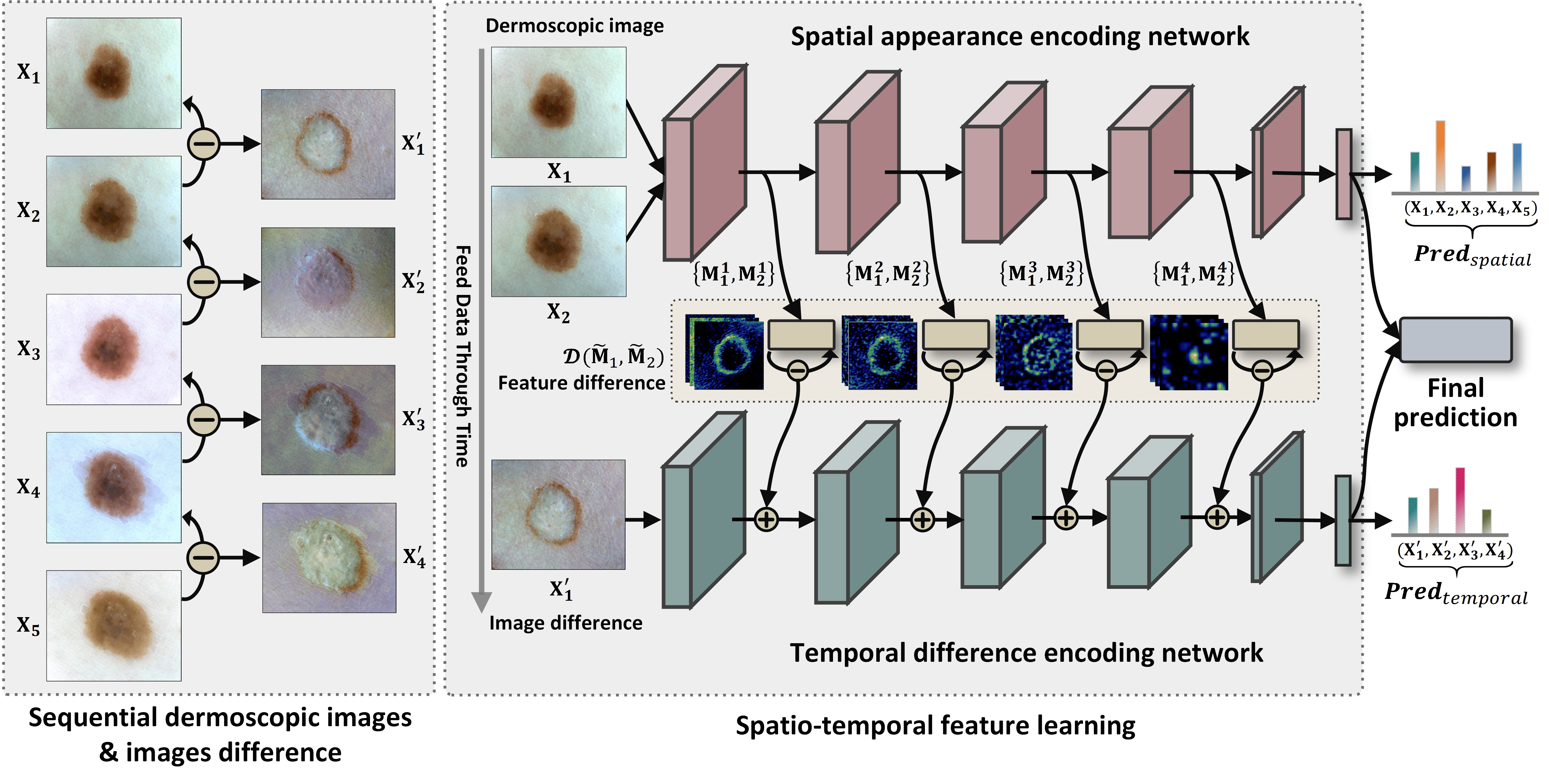}
\caption{The detailed architecture of the proposed framework. The top red blocks consist of the spatial stream network, and the bottom blue blocks denote the temporal stream network. Intermediate brown blocks are online feature subtraction unit which used to extract multi-level features difference} \label{fig2}
\end{figure}

\subsection{Spatial Appearance Encoding Network} The spatial stream network is utilised for encoding dermoscopic images into different levels of appearance abstraction which later can be trained together with temporal features from the temporal network. We employ off-the-shelf ImageNet pre-trained ResNet-34~\cite{ref_11} as the backbone. The last layer of the ResNet-34 was replaced with two linear layers.  We define each image sequence of patient $\textbf{p}$ with $\small{}\textbf{N}$ screenings as $\small{\widetilde{\mbox{\textbf{X}}}^{p} = \left \{ {\textbf{\mbox{X}}_{\scriptsize{\mbox{1}}}^{p}}, 
{\textbf{\mbox{X}}_{\scriptsize{\mbox{2}}}^{p}}, ..., 
{\textbf{\mbox{X}}_{\scriptsize{\mbox{N}}}^{p}} \right\}}$\footnote{$\textbf{p}$ will be omitted in the following sections for clean notation}.
The final decision function of the spatial stream network is defined as follows: 
\begin{equation}
\small{
\textbf{\textit{Pred}}_{spatial} = \delta \left ( \frac{1}{N}\sum_{t=1}^{N}\mathcal{F\left({\mbox{\textbf{X}}}_{\scriptsize{\mbox{\textit{t}}}}, \mbox{\textbf{W}}\right)}\right)}
\end{equation}
where $\scriptsize{\mathcal{F\left({\mbox{\textbf{X}}}_{\scriptsize{\mbox{\textit{t}}}}, \mbox{\textbf{W}}\right)}}$ denotes the mapping function of the spatial encoding sub-network with parameters \small{}\textbf{W} that operates on one dermoscopic image of $\small{\mbox{\textbf{X}}}_{\scriptsize{\mbox{\textit{t}}}}$. We obtain the prediction of spatial appearance encoding network by applying the sigmoid function $\delta$ on averaged outputs from $\small{}\textbf{N}$ screenings. 

\subsection{Temporal difference encoding network} Similar to the spatial network, the temporal network was also implemented using ResNet-34 as the main backbone. Differently, we feed two consecutive images' pixel value difference into the temporal network. Each image difference $\small{{\mbox{\textbf{X}}}'}$ is defined as the pixel-wise value subtraction between two consecutive dermoscopic images. To suppress the noise from irrelevant contexts, we implement the color constancy algorithm $\small{\mathcal{C} \left( \cdot \right)}$ that based on general Gray World \cite{ref_12} and hair removal function  $\small{\mathcal{H} \left( \cdot \right)}$ using morphological filter. 
Concretely, at time $\textit{t}$ we have: 
\begin{equation}
\small
{\textbf{\mbox{X}}}'_{\textit{\mbox{t}}} =  \mathcal{C} \left ( \mathcal{{H}} \left( \mbox{\textbf{X}}_{\mbox{\textit{t}+1}} \right ) \right)
- \mathcal{C} \left ( \mathcal{{H}} \left( \mbox{\textbf{X}}_{\mbox{\textit{t}}} \right)
 \right)
\end{equation}
Our motivation is that subtle dermoscopic changes can be directly reflected by pixels distinction. As shown in Fig. 2, the difference images clearly exhibits the enlargement of the lesion. Thus, we can explicitly learn the temporal evolution of lesions from pixels-level modification.

\subsubsection{Online feature difference extraction:} In contrast to raw pixels, CNN features captured abstract appearance are more robust to image translation and imaging condition changes \cite{ref_14}. Thus, features difference is more suitable for modeling higher-level cross-time dermoscopic distinction. Hence, we further consider incorporating difference information of multi-layer CNN features from spatial encoding network into the temporal sub-network by element-wise feature map addition at the corresponding layers. 

Specifically, in the forward passing of a dermoscopic image sequence, we insert feature subtraction units at each stage of the spatial encoding network to extract multiple levels of features difference among consecutive images:
\begin{equation}
\small
\mathcal{D}\left(  \widetilde{\mbox{\textbf{M}}}_{\scriptsize{\mbox{\textit{t}}}}, \widetilde{{\mbox{\textbf{M}}}}_{\scriptsize{\mbox{\textit{t}+1}}}\right) 
= \left \{  
\left( \mbox{\textbf{M}}_{\scriptsize{\mbox{\textit{t}+1}}}^{1} - \mbox{\textbf{M}}_{\scriptsize{\mbox{\textit{t}}}}^{1}  \right),
\left( \mbox{\textbf{M}}_{\scriptsize{\mbox{\textit{t}+1}}}^{2} - \mbox{\textbf{M}}_{\scriptsize{\mbox{\textit{t}}}}^{2}  \right),\ldots,  
\left( \mbox{\textbf{M}}_{\scriptsize{\mbox{\textit{t}+1}}}^{l} - \mbox{\textbf{M}}_{\scriptsize{\mbox{\textit{t}}}}^{l}  \right) 
\right \}
\end{equation}
where $\small\mbox{\textbf{M}}_{\scriptsize{\mbox{\textit{t}}}}^{l}$ means feature maps of $\small{\mbox{\textbf{X}}}_{\scriptsize{\mbox{\textit{t}}}}$ extracted from $\textit{l}$ layer in the spatial stream network. Hence, the prediction for the temporal stream network is given by:
\begin{equation}
\small{
\textbf{\textit{Pred}}_{temporal} = \delta \left ( \frac{1}{N-1}\sum_{t=1}^{N-1}\mathcal{G \left( {\textbf{\mbox{X}}}'_{\textit{\mbox{t}}}, {\mathcal{D}\left(  \widetilde{\mbox{\textbf{M}}}_{\scriptsize{\mbox{\textit{t}}}}, \widetilde{{\mbox{\textbf{M}}}}_{\scriptsize{\mbox{\textit{t}+1}}}\right)}, {\mbox{\textbf{W}}}'\right)} \right)
}
\end{equation}

Therefore the proposed method can track the dermoscopic changes over time using clues of temporal difference information at both raw pixels-level and abstract features-level instead of only relying on high-level spatial appearance features. 
As for the final prediction, we directly average the output from the two sub-networks before applying sigmoid function. 

\section{Experiment and Results}
We evaluate the performance of the proposed model on a self-collected serial dermoscopic image data with five-fold cross validation. We report both quantitative and qualitative results. Comparison to various baselines and ablation studies are also provided.  
\subsection{Dataset and Implementation}
\subsubsection{Dataset and evaluation:} In this study, we collect 184 serial dermoscopic imaging data a total of 747 dermoscopic images from 184 patients. The dataset is well-balanced and consists of 92 benign lesions and 92 malignant lesions (the malignant lesions include both invasive and in-situ melanoma). Each lesion undergoing digital dermoscopic imaging monitoring was eventually verified by biopsy examination. 
It is worth noting that the length of dermoscopic image sequences varies from 1 to 12, and the average number of images in each image sequence is around 4.06 (details can be seen in the Appendix). In the following experiments, we mainly consider training and testing different models with input sequence length from 2 to 5. 

\subsubsection{Implementation details:} All the experiments were conducted on Pytorch library. The model is optimized using Adam with a batch size of 32 and an initial learning rate of 0.001. During training, we reduce the learning rate by a factor of 5 once the validation loss does not decrease within ten epochs.
During training, we calculate the binary cross-entropy loss separately on the predictions from the two sub-networks and the overall prediction. The weights set for the spatial, temporal and overall network are set as 0.3, 0.3, and 0.4, respectively.
The standard data augmentation techniques such as random resized cropping, colour transformation, and flipping are used in all experiments.  Each dermoscopic input image is resized to the fixed size of 320$\times$320. During the test phase, we utilize ten crops augmentation and then average the final predictions. 

\subsection{Quantitative results.}
We produce four deep learning-based baselines for performance comparison: 1) \textbf{Single-img-CNN:} The Single-img-CNN is trained with lesion images of single time without considering the temporal information; 2) \textbf{CNN-Score-Fusion:} The CNN-Score-Fusion model is similar to the Single-img-CNN, while during test phase we incorporate temporal clue by averaging disease prediction scores of images within input sequence; 3) \textbf{CNN-Feature-Pooling:} The CNN-Feature-Pooling model is directly trained with sequential images by combining CNN features of individual images via average pooling; 4) \textbf{CNN-LSTM:} The CNN-LSTM model, learned on image sequence, performs temporal aggregation over CNN features of sequential dermoscopic images using LSTM.

For a fair comparison, we implement all the models using the same ImageNet pre-trained ResNet-34 as the backbone for CNN feature extraction and adding new learning layers for training. The first three models share similar network architecture which achieved by replacing the last FC layer of the ResNet-34 using two new FC layers with channel number of 32 and a classification layer. The CNN-LSTM is built with two LSTM layers with a hidden size of 32 and a classification layer.  More details can be seen in the Appendix. The evaluation metrics used include accuracy, AUC, precision, sensitivity, and specificity.

\subsubsection{Comparative results:} 
We first compare all the methods at the sequence length of $\small{}\textbf{N}=4$ which is the average length of our sequential dataset. The result is given in Fig. 3 (e) and Table 1. We can see that all the models based on sequential images have better AUC than the Single-img-CNN trained with snapshot images, and the proposed model achieved the best performance with an accuracy of 69.98\%, AUC of 74.34\%, precision of 71.61\%, sensitivity of 69.66\%, and specificity of 70.99\%. This result means the benefit of incorporating temporal clues in melanoma diagnosis, and also demonstrate the effectiveness of the proposed method in leaning spatio-temporal feature from serial images. 

\begin{figure}
\centering
\includegraphics[width=\textwidth]{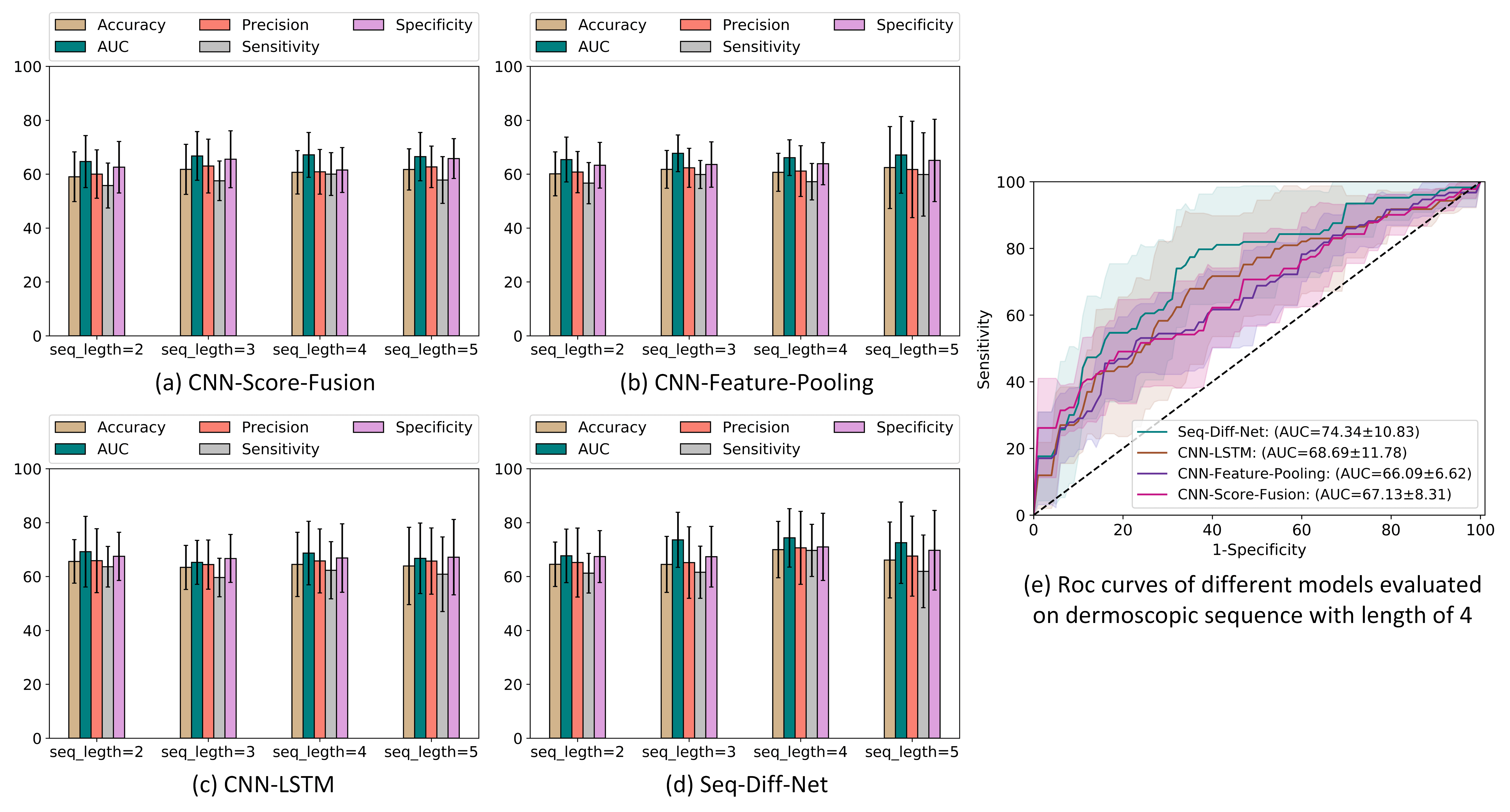}
\caption{(a)-(d): Comparison of the proposed model with other sequence learning models on different lengths of dermoscopic images sequence; (e): Comparision results on the average length of the dataset} \label{fig3}
\end{figure}
\begin{table}
\centering
\caption{The results of comparison study and ablation study. The performance of the sequential models are reported on image sequence with length of 4.}\label{tab1}
\scriptsize{}
{\begin{tabular*}{\textwidth}{@{\extracolsep{\fill} } p{2.7cm}ccccc}
\hline 
Methods & Accuracy(\%) & AUC(\%) & Precision(\%) & Sensitivity(\%) & Specificity(\%) \\
\hline \hline
Single-img-CNN & 61.24$\pm$6.54 & 66.76$\pm$9.87 & 61.51$\pm$8.16 & 58.53$\pm$6.59 & 63.79$\pm$6.39\\
\hline 
CNN-Score-Fusion & 60.67$\pm$8.04 & 67.13$\pm$8.31 & 60.85$\pm$8.29 & 60.02$\pm$ 7.95& 61.53$\pm$8.34\\
CNN-Feature-Pooling & 60.67$\pm$7.05& 66.09$\pm$6.62 & 61.12$\pm$9.39 & 57.21$\pm$6.77 & 63.87$\pm$7.80\\
CNN-LSTM & 64.47$\pm$11.91 & 68.69$\pm$11.78 & 65.76$\pm$11.87 & 62.31$\pm$10.60 & 66.85$\pm$12.69\\
\hline
Our(spatial stream) & 62.89$\pm$5.78 & 68.98$\pm$6.85 & 63.20$\pm$10.47 & 60.45$\pm$5.76 & 65.64$\pm$6.25\\
Our(temporal stream) & 65.54$\pm$7.14 & 70.63$\pm$11.16 & 66.72$\pm$9.51 & 61.85$\pm$6.03 & 69.20$\pm$8.43\\
Our(two-stream) & \bfseries{69.98$\pm$10.48} & \bfseries{74.34$\pm$10.83} & \bfseries{71.61$\pm$13.53} & \bfseries{69.66$\pm$9.68} & \bfseries{70.99$\pm$12.47}\\
\hline
\end{tabular*}}
\end{table}
We then further investigate the temporal reasoning ability of different models by evaluating the performance on different image sequence length from 2$\sim$5. In order to maximum the number of training sequences, we equalise the required input length by padding the first screening or randomly selected consecutive images. The results are shown in Fig. 3 (a)-(d), the point metrics, .i.e, metrics apart from AUC, are obtained on the optimal threshold values. As we can see, there are no obvious performance improvements for all the comparative sequential models when increasing the sequence length. This result leads to the conclusion that score fusion or feature pooling on high-level spatial appearance features only are not good at capturing temporal relation among various observations. By contrast, our proposed model (denoted as Seq-Diff-Net) obtains a consistent AUC boost from 67.72\% to 74.34\% when increasing sequence length from 2 to 4. The AUC drops slightly at the sequence length of 5, which mainly because there are far less $\small{\textbf{N}}=5$ sequence available in the dataset. 

\subsubsection{Ablation results:} We then conduct ablation experiment on the proposed model and summarize the result in Table 1. As it can be seen, the temporal stream network has better performance compared with spatial stream network. The metrics improved $\sim$2.7\%, $\sim$1.7\%, $\sim$3.5\%, $\sim$1.4\%, and $\sim$3.6\% in accuracy, AUC, precision , sensitivity, and specificity, respectively, which demonstrate the effectiveness of incorporating information of pixels difference and CNN features difference for modeling the temporal lesions evolution. Combining the two-stream network, we further gain a large margin of performance improvement.  This result show the superiority of proposed method in diagnosing melanoma using spatio-temporal information from the spatial appearance of individual lesions as well as temporal relations among consecutive lesion images. 

\subsection{Qualitative results.}
\subsubsection{Visualization of temporal difference learning:} 
\begin{figure}
\centering
\includegraphics[width=\textwidth]{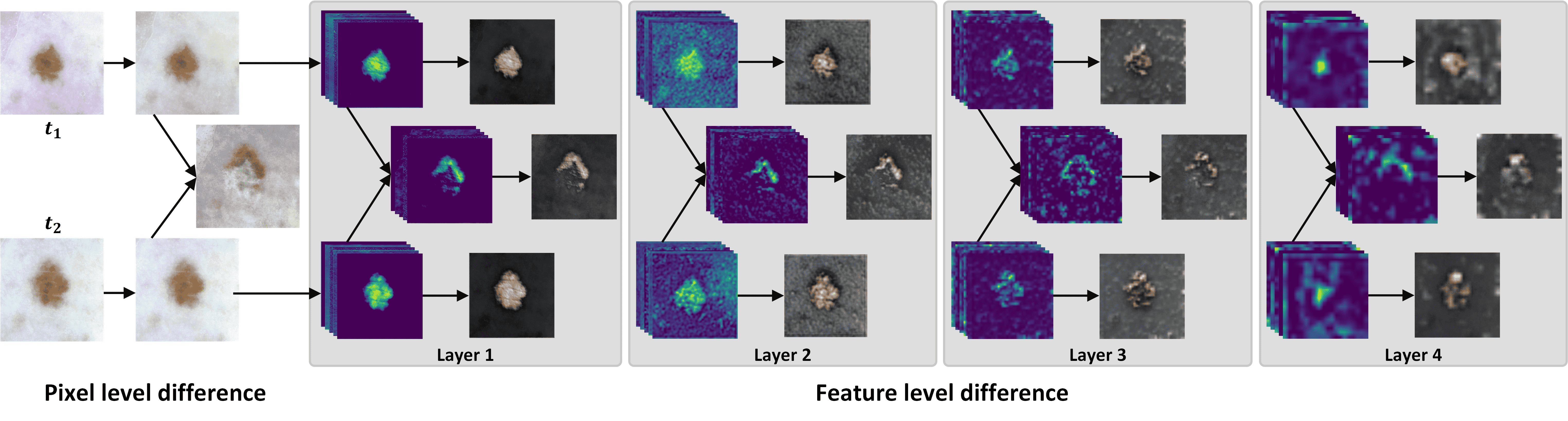}
\caption{Visualisation of the temporal learning process on one case. To intuitively present the visualisation, we first aggregate the feature difference map to RGB space and then further overlay it to the input image.} \label{fig4}
\end{figure}

To better illustrate the temporal difference learning process of the proposed model, we show the pixel-level difference and feature-level difference across various layers on two consecutive images in Fig.~\ref{fig4}. We can observe that the our model has successfully captured the new growth region of the lesion. Moreover, the intermediate convolutional activation maps demonstrates that the final prediction is made in the foreground lesion area. 

\subsubsection{Effect of incorporating temporal information:} To intuitively illustrate how incorporating temporal information affects the decision-making of melanoma diagnosis, we compare prediction results from models that learned with and without sequential dermoscopic images in Table 2. The two malignant melanoma, both show apparent morphological modifications, is successfully detected by the proposed model while misdiagnosed by the model trained with snapshot images. With the adoption of serial images, our model also correctly diagnoses the benign lesion case of patient 2, which exhibits subtle dermoscopic changes across time. By contrast, the Single-img-CNN model still fails in this case due to the lack of temporal clues.

\begin{table}[ht]
\centering
\caption{Effect of incorporating temporal lesion modification information on the decision-making of melanoma diagnosis. Check mark and cross mark denote correct and incorrect diagnostic outcome, respectively.} 
\scriptsize
\begin{tabular}{c|c|c|c|c}
\hline 
 & sequential dermoscopic images & ground truth & Single-img-CNN & Our model \\
\hline\hline
patient 1 & \multicolumn{1}{|m{5cm}|}{\includegraphics[scale=0.054]{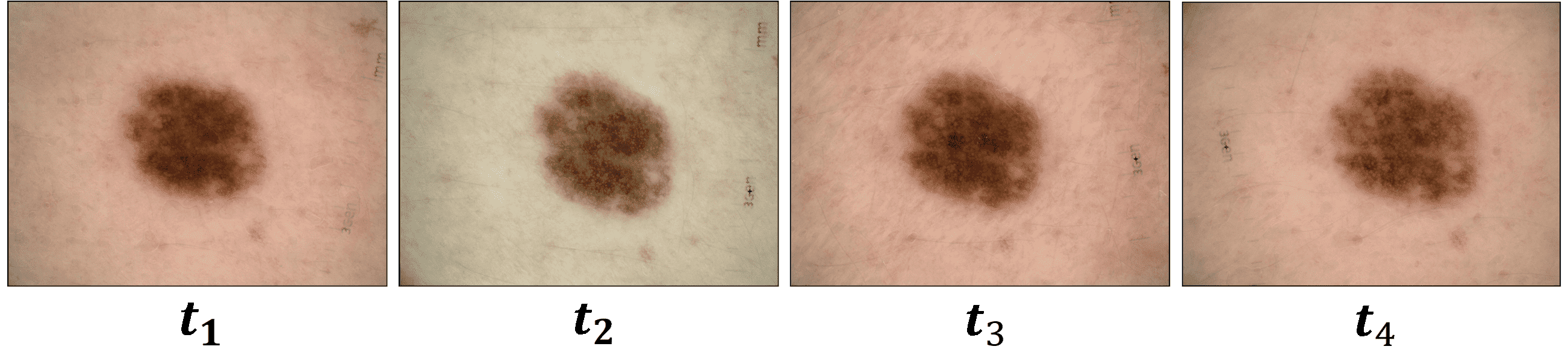}} & benign & \ding[1.5]{51} & \ding[1.5]{51} \\ 
\hline
patient 2 & \multicolumn{1}{|m{5cm}|}{\includegraphics[scale=0.054]{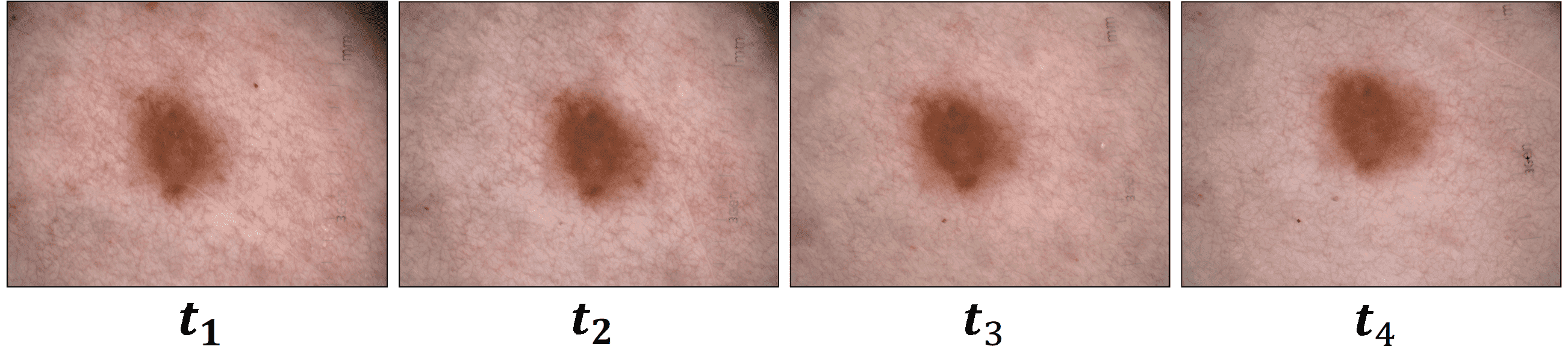}} & benign & \ding[1.5]{55} & \ding[1.5]{51} \\
\hline
patient 3 & \multicolumn{1}{|m{5cm}|}{\includegraphics[scale=0.054]{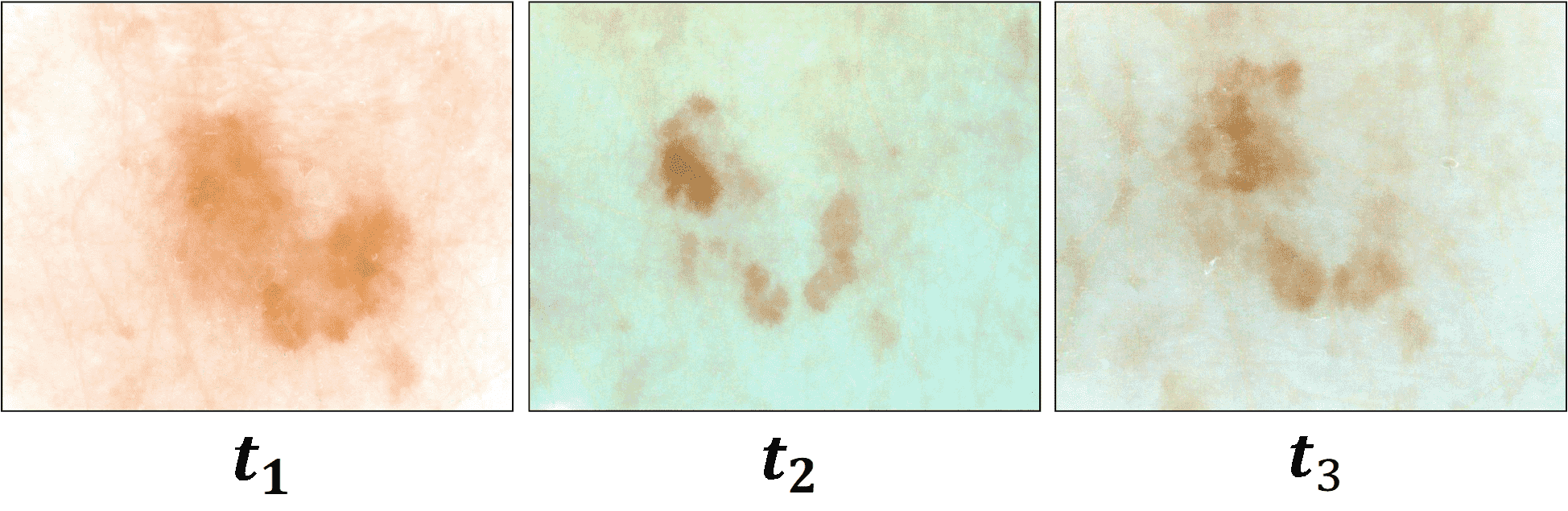}}  & malignant & \ding[1.5]{55} & \ding[1.5]{51} \\
\hline
patient 4 & \multicolumn{1}{|m{5cm}|}{\includegraphics[scale=0.054]{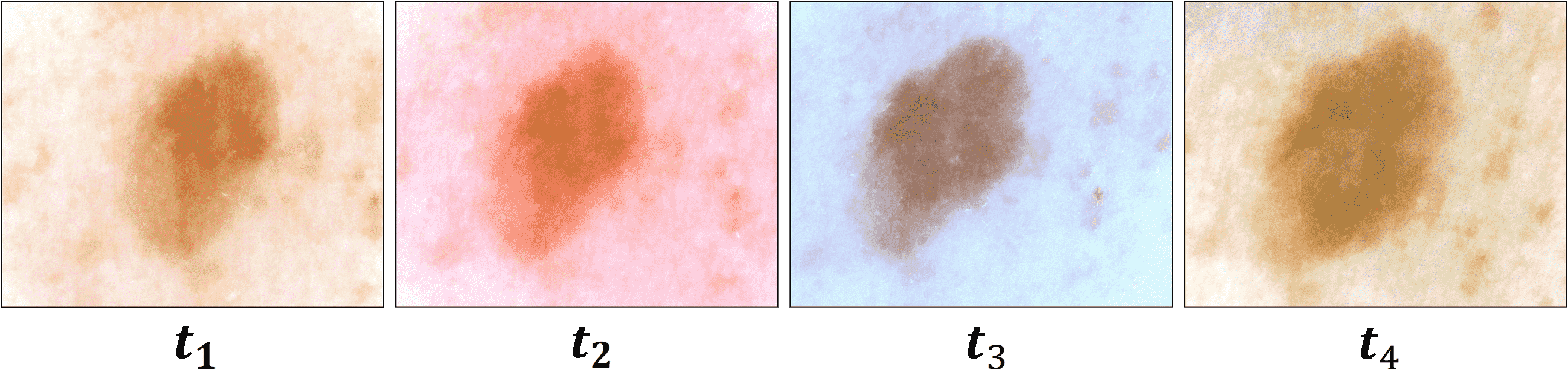}} & malignant & \ding[1.5]{55} & \ding[1.5]{51} \\
\hline
\end{tabular}
\end{table}
\section{Conclusion}
In this study, we propose to model the spatio-temporal features of a lesions growth from sequential dermoscopic images for automated melanoma diagnosis. We achieve this by constructing a two-stream network architecture which is capable of simultaneously learning dermoscopic appearance features from individual lesions while explicitly capturing temporal relations among consecutive images of lesions from both pixels difference and multi-level CNN features difference. Experiments on our serial dermoscopic image data show our method can achieve significant performance improvement than that of only using dermoscopic images from just a single time point, as well as other commonly used sequence learning methods.

\bibliographystyle{splncs04}

\newpage
\section{Appendix}

\subsubsection{Configuration of different models:}
We give detailed architecture configurations of the proposed two-stream network and the other comparative models in Table 3 and Fig. 5. As aforementioned, the spatial stream network and the temporal network of the proposed method share similar network architecture, also, the the Single-img-CNN, CNN-Score-Fusion, CNN-Feature-Pooling have same network configuration.

\begin{figure}
\centering
\includegraphics[width=\textwidth]{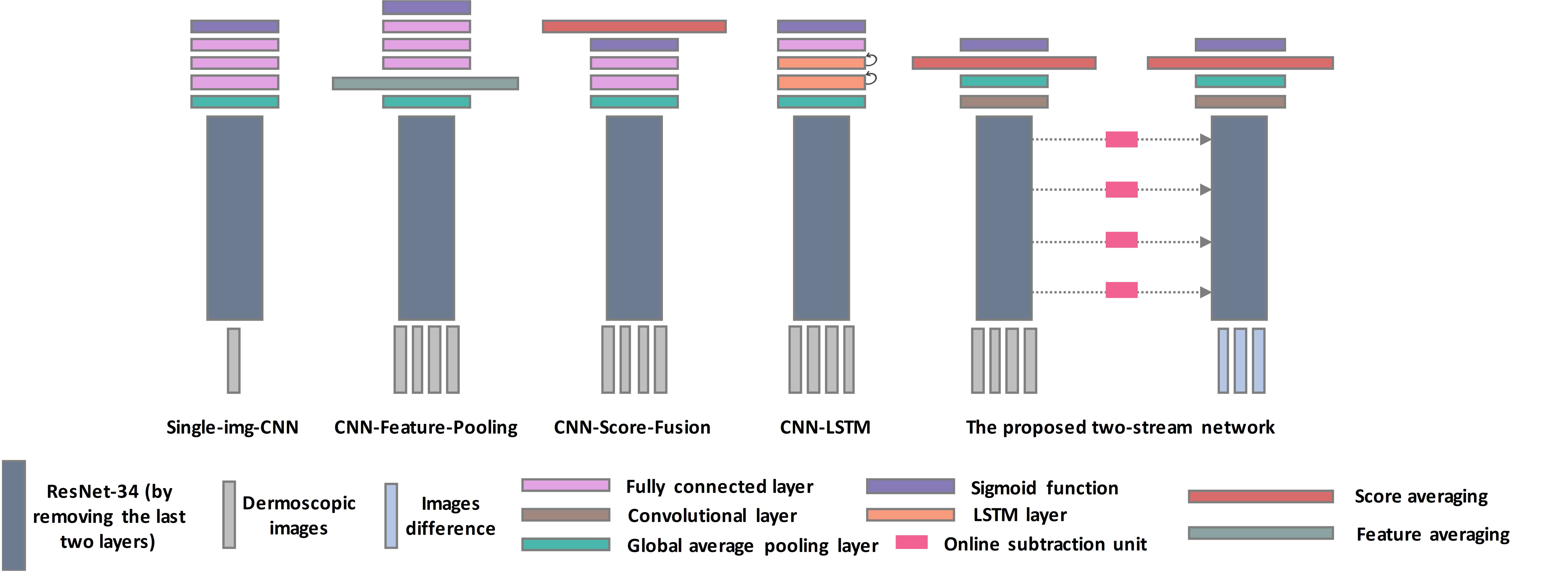}
\caption{The architectures of the proposed model and other comparative models. We do not show the dropout layer, batch normalization layer and activation layer in the figure for simplicity.} 
\end{figure}\label{fig5}

\begin{table}
\centering
\caption{Configuration of the proposed model and the other comparative models.Conv, BN, ReLU denotes convolutionaly layer, batch normalization layer and activation function of rectified linear unit, respectively.}
\scriptsize
\begin{tabular}{c|c|c}
\hline
\begin{tabular}[c]{@{}c@{}}\textbf{Single-img-CNN}\\ \textbf{CNN-Score-Fusion}\\ \textbf{CNN-Feature-Pooling}\end{tabular} & \textbf{CNN-LSTM} & \begin{tabular}[c]{@{}c@{}}\textbf{Spatial stream network} \textbf{\&}\\ \textbf{Temporal stream network}\\ \textbf{of the two-stream nework}\end{tabular} \\ \hline \hline
\multicolumn{3}{c}{\begin{tabular}[c]{@{}c@{}}ResNet 34 by removing the last global averaging layer and \\ the fully connected layer\end{tabular}} \\ \hline 
\multicolumn{2}{c|}{Global average pooling} & Dropout, p=0.5 \\ \hline
\multicolumn{2}{c|}{Dropout, p=0.5} & Global average pooling \\ \hline
FC, 512$\times$32 & \begin{tabular}[c]{@{}c@{}}LSTM layer, hidden size=32\\ Dropout, p=0.5\end{tabular} & Dropout, p=0.5 \\ \hline
\begin{tabular}[c]{@{}c@{}}Dropout, p=0.5\\  BN \& ReLU\end{tabular} & \begin{tabular}[c]{@{}c@{}}LSTM layer, hidden size=32\\ Dropout, p=0.5\end{tabular} & \begin{tabular}[c]{@{}c@{}}FC, 512$\times$16\\ BN \& ReLu\end{tabular} \\ \hline
FC, 32$\times$32 & \multirow{4}{*}{FC, 32$\times$1} &  \multirow{4}{*}{FC, 16$\times$1} \\ 
\cline{1-1} \begin{tabular}[c]{@{}c@{}}Dropout, p=0.5\\ BN \& ReLU\end{tabular} &  & \\
\cline{1-1} FC, 32$\times$1 &  &  \\ \hline
\multicolumn{3}{c}{Sigmoid function} \\ \hline
\end{tabular}
\end{table}

\newpage
\begin{figure}
\centering
\includegraphics[width=\textwidth]{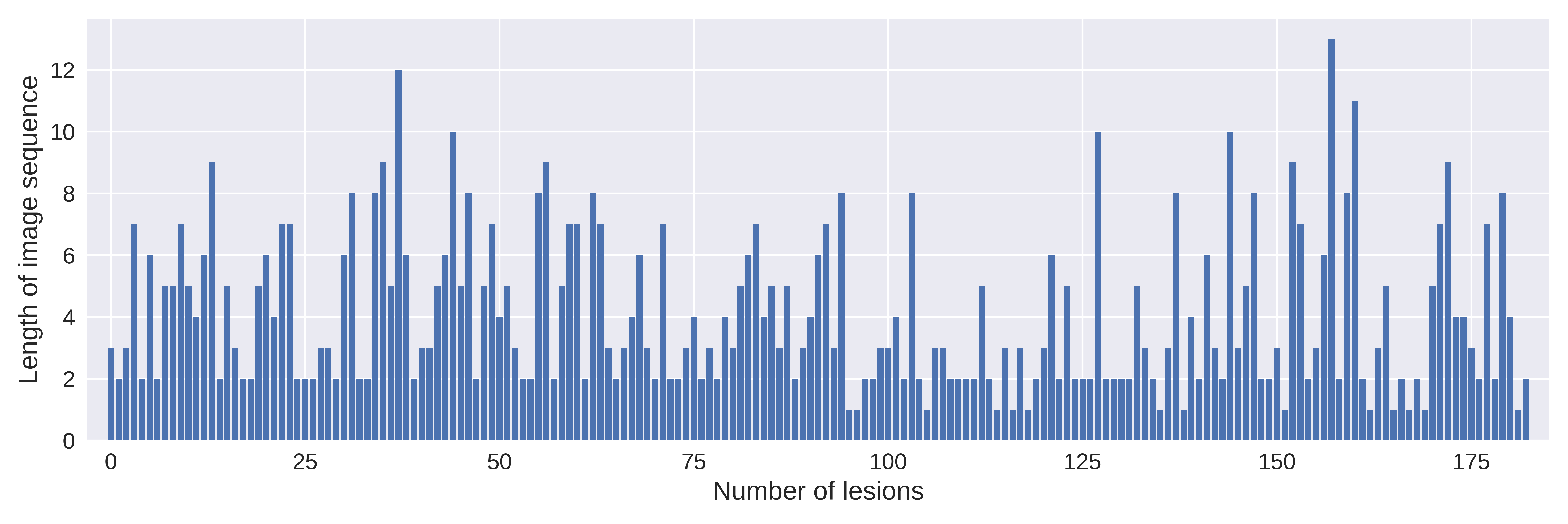}
\caption{The length distribution of dermoscopic images sequence in our collected dataset.} 
\end{figure}\label{fig6}

\end{document}